\documentclass[journal]{IEEEtran}
\usepackage{amsmath,amsfonts}
\usepackage{algorithmic}
\usepackage{algorithm}
\usepackage{array}
\usepackage[caption=false,font=normalsize,labelfont=sf,textfont=sf]{subfig}
\usepackage{textcomp}
\usepackage{stfloats}
\usepackage{url}
\usepackage{verbatim}
\usepackage{graphicx}
\usepackage{cite}

\usepackage{hyperref}
\usepackage[para]{footmisc}
\usepackage[inline]{enumitem}
\usepackage{booktabs}

\usepackage{amssymb}

\usepackage{capt-of}
\usepackage{xcolor}
\usepackage[most]{tcolorbox}
\usepackage{etoolbox}

\usepackage{listings}
\newcounter{listing}
\renewcommand{\thelisting}{\arabic{listing}} 

\definecolor{codebackground}{rgb}{0.97, 0.97, 0.97}
\definecolor{keywordpurple}{rgb}{0.85, 0.20, 0.50}
\definecolor{codegreen}{rgb}{0,0.6,0} 
\definecolor{codegray}{rgb}{0.1,0.1,0.1} 
\definecolor{codepink}{rgb}{0.85, 0.20, 0.50}
\definecolor{backcolour}{rgb}{0.95,0.95,0.92}
\usepackage{inconsolata} 
\lstdefinestyle{pythonstyle}{
    basicstyle=\ttfamily\footnotesize\color{codegray},
    commentstyle=\color{codegreen},
    keywordstyle=\color{codepink},
    stringstyle=\color{keywordpurple},
    breaklines=true,
    showstringspaces=false,
    frame=none,
    columns=fixed,
    basewidth=0.52em,
    lineskip=-0.4pt,
    aboveskip=5pt,
    belowskip=5pt
}
\newtcblisting{pythoncode}{
    arc=15pt,                
    boxrule=0pt,
    colback=codebackground,
    colframe=codebackground,
    listing only,
    listing engine=listings,
    listing options={
            style=pythonstyle,
            language=Python,
            alsodigit={.:},
        },
    enhanced,
    boxsep=0pt,  
    left=10pt,  
}

\begin{document}

\title{A Unified Experimental Architecture for Informative Path Planning: from Simulation to Deployment with \textit{GuadalPlanner}}

\author{
Alejandro Mendoza Barrionuevo, Dame Seck Diop, Alejandro Casado Pérez, Daniel Gutiérrez Reina and Sergio L. Toral Marín, Samuel Yanes Luis

\thanks{The authors are with the Department of Electronic Engineering, University of Sevilla, Spain (e-mail: {\tt\footnotesize amendoza1@us.es}; {\tt\footnotesize dseck@us.es}; {\tt\footnotesize acasado4@us.es}; {\tt\footnotesize dgutierrezreina@us.es}; {\tt\footnotesize storal@us.es}; {\tt\footnotesize syanes@us.es})
}
}

\maketitle

\begin{abstract}
The evaluation of informative path planning algorithms for autonomous vehicles is often hindered by fragmented execution pipelines and limited transferability between simulation and real-world deployment. This paper introduces a unified architecture that decouples high-level decision-making from vehicle-specific control, enabling algorithms to be evaluated consistently across different abstraction levels without modification. 
The proposed architecture is realized through GuadalPlanner, which defines standardized interfaces between planning, sensing, and vehicle execution. It is an open and extensible research tool that supports discrete graph-based environments and interchangeable planning strategies, and is built upon widely adopted robotics technologies, including ROS2, MAVLink, and MQTT.
Its design allows the same algorithmic logic to be deployed in fully simulated environments, software-in-the-loop configurations, and physical autonomous vehicles using an identical execution pipeline. The approach is validated through a set of experiments, including real-world deployment on an autonomous surface vehicle performing water quality monitoring with real-time sensor feedback.
\end{abstract}

\IEEEpeerreviewmaketitle

\section{Introduction}

\IEEEPARstart{I}{nformative} Path Planning (IPP) is a central topic in autonomous robotics, where the objective is not simply to reach a destination, but to decide where to go in order to maximize the information gathered about an environment. These algorithms are particularly relevant in scenarios such as environmental monitoring, pollution mapping, infrastructure inspection or search and rescue, where data acquisition efficiency is critical. IPP algorithms often involve probabilistic models, online decision-making, and real-time sensor feedback, which makes their development and testing inherently complex.

Despite extensive research on IPP strategies—from greedy heuristics to deep reinforcement learning \cite{khan2014greedy,jaraten2024aquafel,yanes2024deep,barrionuevo2025optimizing}—most algorithms are usually only validated in highly controlled, custom-built simulation environments. These simulators typically focus on the specific needs of the research group and lack generality, making them unsuitable for cross-validation or code reuse. Moreover, the absence of standardized tools capable of bridging the gap between simulation and real-world deployment severely limits practical applicability. In particular, very few systems allow testing the same algorithm transparently in both virtual and physical platforms without substantial reconfiguration or rewriting. 
While numerous robotic simulators exist \cite{craighead2007survey, mualla2018comparison}, they are primarily designed for low-level dynamics, perception, or control validation, and do not directly support the experimental requirements of IPP research. 
If the decision-making layer is decoupled from execution and sensing, then IPP algorithms can be evaluated reproducibly across simulation and real deployments without modification.

To the best of the authors’ knowledge, we introduce \textit{GuadalPlanner} as the first end-to-end experimental platform explicitly designed to bridge this gap. This is a modular and extensible experimental architecture for the development, testing and deployment of IPP algorithms in autonomous vehicles.
The system provides an appropriate separation between high-level decision-making and low-level execution, enforcing standardized interfaces for navigation, sensing, and communication.
A core design feature of GuadalPlanner is its interface consistency: planning algorithms interact with simulated and real vehicles identically. 
In GuadalPlanner, IPP algorithms interact exclusively with a unified fleet interface, while motion execution and sensing are delegated to interchangeable backends.

We propose a execution model organized into three abstraction levels: 
\begin{enumerate*}[label=(\roman*)]
    \item Algorithm-level simulation, for prototyping using ideal vehicle models in graph-based environments;
    \item Software-in-the-Loop, which emulates realistic control behavior with full middleware integration;
    \item Real deployment, where the same code runs on physical vehicles equipped with 
    autopilot, navigation sensors, and wireless communication.
\end{enumerate*}

For integration with physical vehicles, GuadalPlanner is designed around three open-source and widely adopted technologies: ArduPilot\footnote{\url{https://ardupilot.org/}}, ROS2\footnote{\url{https://www.ros.org/}}, and MQTT\footnote{\url{https://mqtt.org/}}. Each of these components plays a key role in the architecture. 
At the core of vehicle control, ArduPilot is a mature and widely adopted autopilot software supporting aerial, ground, and aquatic platforms. It includes a Software-in-the-Loop (SITL) simulator that replicates vehicle dynamics, control, and MAVLink messaging protocol communication, enabling realistic mission testing with the same software used in real deployments. ROS2 acts as middleware in GuadalPlanner, providing modular and reusable components for planning, communication, and control. Through MAVROS, it bridges ROS2 messages with MAVLink, enabling integration with ArduPilot-based systems. Communication between decision modules and vehicles—simulated or real—is handled via MQTT, allowing both local and remote execution, which is useful for multi-agent scenarios or low-resource onboard systems. 

Given this architectural design, GuadalPlanner is currently compatible with a wide range of autonomous vehicles equipped with a MAVLink-based autopilot and whose motion can be modeled in a two-dimensional workspace.
The system does not impose assumptions on the vehicle type or sensing payload, as these aspects are encapsulated within interchangeable interface layers, but it relies on a planar, graph-based environment representation, which makes it particularly well suited for ground vehicles, surface vessels, or aerial vehicles operating at a fixed altitude.
All planning, sensing, and decision-making processes are defined over this 2D abstraction, while vehicle-specific actuation and control are handled at lower layers. 

This architecture supports the complete IPP development cycle from design to field execution without software changes. Researchers can progressively increase the realism and complexity of their experiments while reusing the same codebase and interface definitions. 
One of the main goals of this work is to provide researchers with a practical and reusable tool for developing and deploying IPP algorithms. 
To this end, we first introduce a design proposal that aims to unify control interface and to abstract decision-making from vehicle-specific execution. 
GuadalPlanner is the open-source implementation of this proposal and serves as its experimental validation.
It also encourages reproducibility, since standardized modules and environments can be shared across research groups. 
Multi-agent configurations are supported, environment modularization, and integration of real-time sensor feedback, making it a versatile tool for a wide range of IPP applications.

To validate the architecture, a series of standard experiments for IPP were conducted across the three defined abstraction levels: fully simulated, SITL, and real-world deployment with an actual fleet of ASVs. 
These experiments serve as a proof-of-concept on how GuadalPlanner enables systematic evaluation of IPP strategies. They provide a bridge between algorithmic design and practical deployment. 
Real-world implementation was carried out using an autonomous surface vehicle, executing algorithms previously tested at higher levels of abstraction. Several IPP algorithms from the literature were implemented and evaluated, including greedy exploration based on uncertainty \cite{khan2014greedy}, expected improvement using Bayesian optimization \cite{peralta2023water}, and systematic coverage strategies \cite{ghasemi2024flood}. 
By testing in environments with diverse characteristics—both synthetic and real—these experiments demonstrate the system’s ability to support different planning strategies and mission objectives, while maintaining identical execution pipelines across simulation and physical platforms.

In summary, the main contributions of this work are as follows: 
\begin{itemize}
    \item The proposal of a unified experimental architecture for IPP which, to the best of the authors’ knowledge, constitutes the first one that decouples high-level decision-making from vehicle-specific control. It is specifically designed to support end-to-end IPP experimentation across simulation and it is built on established robotics technologies such as ROS2, MAVLink, and MQTT. This abstraction enables the same algorithm logic to be executed across simulated, software-in-the-loop, and real-world platforms without modification. An experimental validation has also been carried out, including a real ASV implementation.

    \item The introduction of an open and extensible research tool that supports systematic and comparative evaluation of IPP algorithms, facilitating reproducible experimentation and narrowing the gap between algorithm development, simulation-based testing, and field deployment.
\end{itemize}

\section{Related Work}

\begin{table*}[t] 
\centering 
\caption{Summary of IPP works, architectures employed, and vehicle types.} \label{tab:algorithms_comparative} 
\scriptsize 
\begin{tabular}{cp{5cm}p{4.5cm}ccc} 
\toprule \textbf{Ref} & \textbf{IPP Approach} & \textbf{Experimental Architecture} & \textbf{Multi-agent} & \textbf{Vehicle Type} & \textbf{Real Impl.} \\ 
\midrule

\cite{IPPalrashdia} & Oil spill estimation using Bayesian Optimization with image-based samples. & Custom implementation based on image processing. & Yes & UAV/ASV & No \\
\midrule
 
\cite{IPPaltahat} & 3D navigation using Genetic Algorithms to find underwater high-polluted areas. & Custom Java environment. & No & AUV & No \\  
 \midrule

\cite{IPPangelyn} & Coastal monitoring using Proximal Policy Optimization for coverage and detection. & Custom Python implementation using Gymnasium. & No & AUV & No \\  
 \midrule

\cite{IPPdutta} & Active regression via Mixed Integer Programming for provably optimal solutions. & Custom simulation with GUROBI solver. & Yes & AUV
 & No \\  
\midrule

\cite{IPPkailas} & Spatiotemporal prediction using Gaussian Processes and submodular function optimization. & Custom multi-agent simulation with real-world datasets. & Yes & UAV
 & No \\  
\midrule

\cite{IPPlindgren} & Area monitoring against
intruders via Receding Horizon Optimization. & Custom Matlab grid-based simulations. & Yes & UGV & No \\  
\midrule

\cite{IPPruckin} & Autonomous acquisition of
informative training images using deep learning techniques. & Custom Python implementation. & No & UAV & No \\  
\midrule

\cite{IPPwiman} & Real-time surveillance for intruder detection using the Autonomous Surveillance Planner. & Custom C++ implementation, tested in ROS framework with Gazebo simulator. & No & UGV & Yes \\  
\midrule

\cite{IPPyu} & Ocean parameter sampling via hybrid Q-learning and Probabilistic Roadmap. & Custom Python simulation with public reanalysis data. & No & AUV & No \\  
\midrule

\cite{IPPyunze} & Parking occupancy estimation solved with Monte Carlo Bayes Filter Tree. & Custom Matlab Automated Parking Valet toolbox simulations. & No & UGV & No \\  
\midrule

\cite{he2019autonomous} & Chemical leak mapping using a modified Potential Field algorithm. & Custom platform with ROS field tests. & Yes & UAV & Yes
 \\  
\midrule

\cite{jaraten2024aquafel} & Water quality monitoring via AquaFeL-PSO and Federated Learning. & Custom Python simulation using Scikit-learn and DEAP. & Yes & ASV
 & No \\  
\midrule

\cite{khan2014greedy} & Maximizing value of information in underwater networks using a Greedy path planner. & Custom implementation for grid-based simulations. & No & AUV
 & No \\  
\midrule

\cite{peralta2023water} & Multi-objective online modeling using Bayesian Optimization and Voronoi region partitioning. & Custom Python implementation. & Yes & ASV
 & No \\
\midrule

\cite{yanes2024deep} & Water quality monitoring using jointly local Gaussian processes and deep reinforcement learning. & Custom Python implementation. & Yes & ASV
 & No \\
 
\bottomrule 
\end{tabular} 
\end{table*}

Informative Path Planning has been extensively studied across a wide range of application domains, including environmental monitoring, surveillance, infrastructure inspection, and data collection. Prior work has proposed diverse planning strategies based on greedy heuristics \cite{khan2014greedy}, Bayesian optimization \cite{peralta2023water, IPPalrashdia}, particle swarm optimization \cite{jaraten2024aquafel}, evolutionary methods \cite{IPPaltahat}, mixed-integer programming \cite{IPPdutta}, and deep reinforcement learning (DRL) \cite{IPPangelyn, yanes2024deep}. These algorithms aim to maximize information gain while minimizing travel cost, adapting online to the environment as new data is acquired. 
They have demonstrated strong performance in specific scenarios and tasks using unmanned aerial vehicles (UAVs), autonomous underwater vehicles (AUVs), ASVs, or unmanned ground vehicles (UGVs), among others, as summarized in Table~\ref{tab:algorithms_comparative}. 
However, despite the rich literature on IPP, there is a notable lack of standardized simulation and deployment architecture that allow for fair and reproducible experiments between different algorithms. 
As shown in Table~\ref{tab:algorithms_comparative}, most studies validate their methods on custom-built simulators or specific testbeds, often tightly coupled to a specific vehicle model, sensing modality, or task definition. Examples include Matlab-based simulations \cite{IPPlindgren, IPPyunze}, Python environments \cite{IPPruckin, IPPkailas, yanes2024deep, peralta2023water}, or Java-based simulators \cite{IPPaltahat}. It is also usual to find that plenty of these previous works lack of a public repository of code, as noted by \cite{POPOVIC2024104727}.
While these implementations are sufficient for validating individual algorithms, they significantly limit reproducibility and the practical transferability of the proposed solutions. Indeed, most of the related work approaches do not provide tools that can bridge the gap between simulation and real-world deployment.

\begin{figure*}[htbp]
    \centering
    \includegraphics[width=0.9\textwidth]{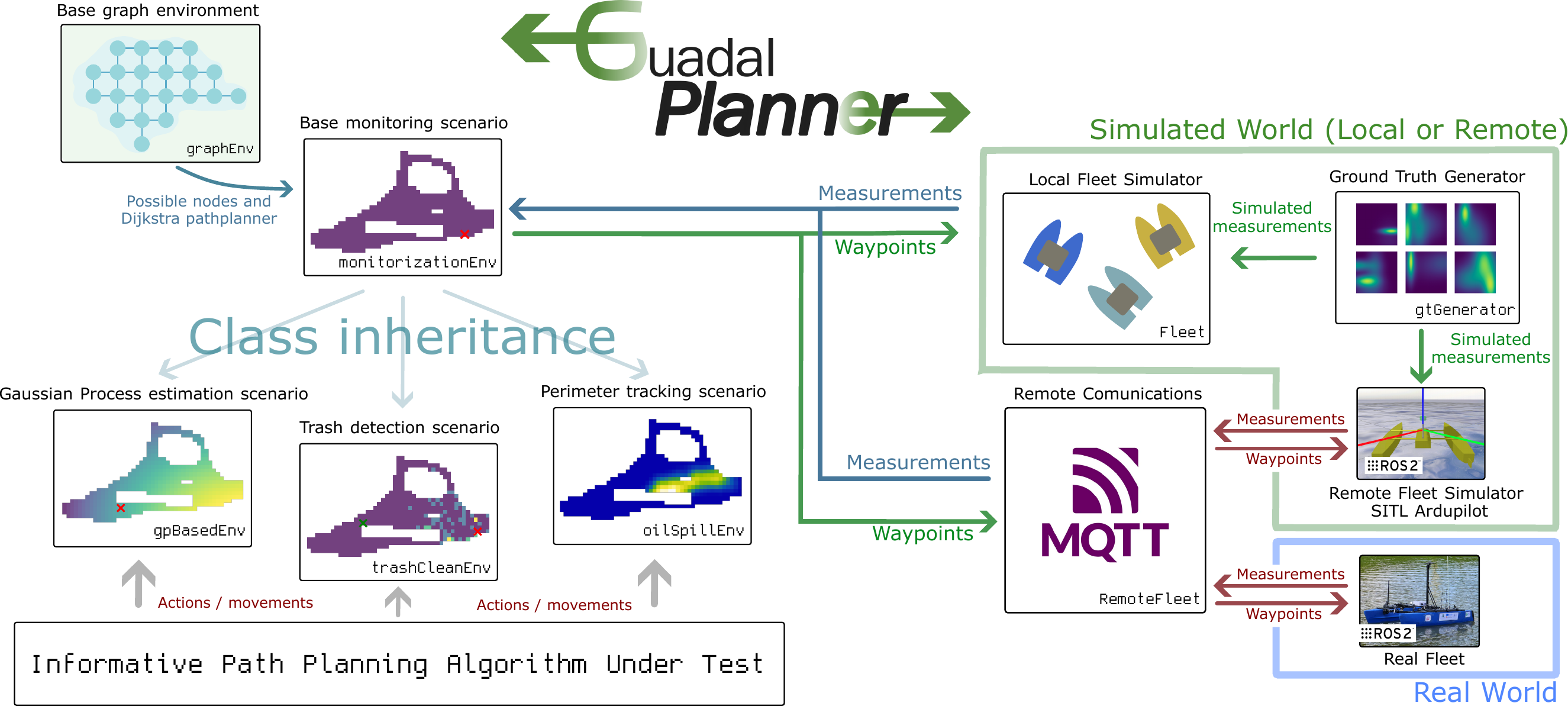}
    \caption{Diagram of the GuadalPlanner implementation. It is structured around a hierarchical architecture of classes that progresses from general to specific.}
    \label{fig:GuadalPlannerScheme}
\end{figure*}

A smaller subset of works integrates IPP algorithms within robotics middleware or robotic platforms. Some studies employ ROS and Gazebo for simulation or field testing \cite{IPPwiman, he2019autonomous}, while others validate algorithms using real datasets \cite{IPPyu} or partial hardware deployments. However, even in these cases, the planning logic is often intertwined with vehicle-specific execution layers. The transition from simulation to real world deployment requires non-trivial redesign of the pipeline, which is rarely preserved across abstraction levels and is usually tailored for a specific kind of fleet.

Regarding the multi-agent capabilities of previous IPP frameworks, Multi-agent IPP is relevant among previous approaches like \cite{IPPalrashdia, IPPkailas, peralta2023water}, but these focus on scenarios specifically designed for the study, without the possibility of exchanging their modules, many of them with static simulation settings.
In these works, computational complexity and fleet scalability is evaluated algorithmically, but not architecturally: adding vehicles or changing platforms usually implies redesigning the software stack. 
The gap is severe given the growing interest in civil applications of unmanned vehicles in a variety of fields, such as autonomous vehicles in urban traffic scenarios \cite{figueiredo2009simulate}, environmental monitoring in remote or hazardous areas \cite{he2019autonomous} or precision agriculture \cite{liu2022challenges}, among others. 

Several works have analyzed and employed high-fidelity simulation platforms for autonomous vehicle research, but they often focus on low-level simulation of vehicles \cite{craighead2007survey}. In \cite{mualla2018comparison}, a comparative study of agent-based UAV simulation frameworks identifies Gazebo and AirSim as the most suitable tools in terms of realism and extensibility. These platforms are widely adopted in robotics due to their accurate physics engines and rich graphical environments. Similarly, \cite{batista2024deep} leverages graphically complex simulators such as Isaac Sim and Gazebo to train deep reinforcement learning policies for low-level control of autonomous surface vehicles, where the learning agent directly commands individual thrusters.
Other works have explored SITL-based validation pipelines to bridge the simulation-to-real gap. For instance, \cite{kumar2019simulation} employs DroneKit SITL to coordinate multiple UAVs in cooperative missions. Although this approach improves realism at the control level, the planning and execution layers remain closely intertwined and tailored to specific vehicle configurations.
In general, these approaches focus on the low-level fidelity of the vehicle dynamics, but neglect the environment modeling, which is a cornerstone of any IPP problem. Some works are useful for control problems, and others for abstract/general IPP. Surprisingly, few of them seem to address the full stack of simulation that is required for a simulation-to-reality deployment.

In this sense, while existing simulators are well suited for low-level control, perception, and dynamics modeling, they are not designed to support systematic evaluation of high-level IPP strategies. In most approaches, planning logic is embedded within simulator-specific execution loop, tightly coupling decision-making with vehicle control and environment representation. 
Overall, existing IPP literature emphasizes algorithmic innovation over experimental unification. 
As a result, to the best of our knowledge, there is currently no widely adopted architecture that decouples high-level IPP decision-making from vehicle-specific control while supporting consistent validation across simulation and real-world deployment. 
This limitation hinders comparative evaluation, slows experimental iteration, and increases the gap between algorithm development and field deployment.

GuadalPlanner is proposed to address this research gap by providing a unified experimental architecture that standardizes the interaction between IPP algorithms, environment models, and autonomous vehicle controllers. Rather than introducing a new planning method, the contribution focuses on enabling reproducible, comparable, and transferable IPP experimentation across abstraction levels and platforms, leveraging established robotics tools while preserving algorithmic flexibility.

\section{Methodology}
This section introduces the unified execution abstraction architecture for IPP, aimed at decoupling high-level decision-making from vehicle-specific execution. The proposed methodology formalizes the interaction between planning logic and vehicle control, enabling IPP algorithms to operate independently of the underlying platform while preserving identical execution semantics across simulation and real-world deployment. By establishing functional equivalence between these execution contexts, the approach supports reproducible experimentation and systematic comparison of IPP strategies under increasing levels of realism.

GuadalPlanner implements this abstraction as a modular and extensible experimental architecture. In field conditions, it supports a closed-loop planning and execution process designed for MAVLink-compatible autonomous vehicles, and relies on a middleware layer designed around ROS2 for coordination and communication.

The proposed execution abstraction is structured around three deployment levels with increasing execution fidelity, each addressing a different stage of IPP algorithm evaluation. This layered design enables systematic testing of decision-making strategies under progressively realistic conditions while preserving identical planning logic across all levels:

\begin{enumerate}
    \item \textbf{Algorithm-level simulation:} This level focuses on the validation of high-level IPP logic in abstract, discrete environments. Simplified vehicle models and synthetic sensing are used to isolate decision-making behavior from physical constraints, enabling rapid prototyping, debugging, and comparative evaluation of planning strategies.
    
    \item \textbf{Software-in-the-Loop (SITL):} At this level, realistic vehicle dynamics and autopilot behavior are introduced through ArduPilot SITL, while maintaining the same communication and execution interfaces. This stage bridges the gap between abstract simulation and real deployment, allowing planners to be evaluated under realistic motion constraints without modifying their internal logic.
    
    \item \textbf{Real deployment:} The final level executes the same planning algorithms on physical autonomous vehicles operating in real environments, integrating live sensor data and real-time actuation. This level enables the transferability of IPP strategies from simulation to field conditions without architectural changes.
\end{enumerate}

By maintaining functional equivalence across these levels, GuadalPlanner supports the development of reproducible algorithms that can be progressively tested and deployed without architectural rewrites. The formalization of a repeatable experimental pipeline for IPP empowers researchers to focus on innovation rather than infrastructure.

From a structural perspective, the architecture follows an internal hierarchical abstraction model, where increasingly specialized components extend a shared execution interface. This organization promotes systematic reuse of modules such as planning logic, environments, or vehicles, among others.
Accordingly, GuadalPlanner is designed using class inheritance capabilities of the object-oriented Python programming language. From the most abstract scenario to the most specific, the software defines the proper level of fidelity for validation, as illustrated in Fig.~\ref{fig:GuadalPlannerScheme}. 
One of the central classes is \textit{Fleet}, which is responsible for managing fleets of vehicles in local simulation. For scenarios in which a fleet of real or distributed vehicles is controlled, there is the \textit{RemoteFleet} class, which inherits from \textit{Fleet} and extends its functionality to include remote communication. Both classes operate on a discrete, graph-based environment for path planning, represented by the \textit{GraphEnv} class. It models the navigation space and allows the definition of constraints, positions, and feasible routes.

Based on this representation, higher-level abstractions are incorporated into \textit{MonitorizationEnv}, which inherits basic properties and introduces navigation logic, sensor modeling, and communication mechanisms to the environment. 
These components serve as templates for defining application-specific scenarios, such as environmental model estimation, target localization (e.g., floating debris), or oil spill tracking. In addition, the architecture naturally supports multi-vehicle configurations, enabling the implementation of coordination strategies including collaborative behavior, task distribution, or conflict resolution in shared environments.
Performance evaluation are usually scenario-dependent, as the definition of meaningful metrics strongly depends on the specific IPP objective. Thus, proposed architecture allows users to define task-specific metrics aligned with their particular IPP objectives. At the same time, a minimal set of general-purpose evaluation metrics is provided and computed across scenarios, including traveled distance per vehicle, mission duration, or uncertainty-related indicators, among others. However, some metrics are only calculated when applicable, such in the case of the root mean square error with respect to the ground truth, calculated in simulation as it is not available in the real world.

\begin{figure}[t]
    \centering
    \includegraphics[width=0.48\textwidth]{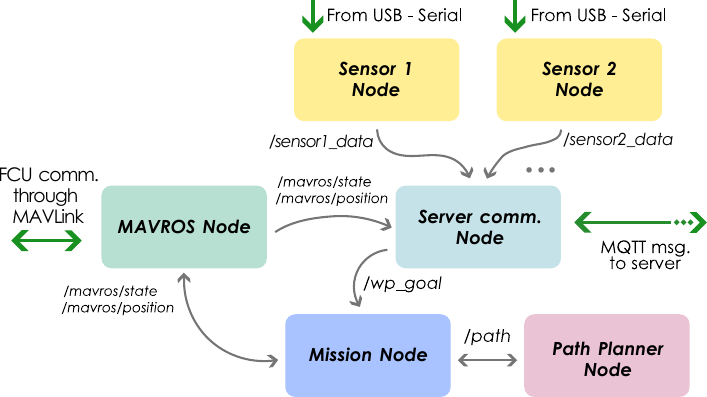}
    \caption{Internal structure of the proposed middleware architecture, illustrating the interaction between planning, communication, and vehicle interfaces implemented over ROS2. The green arrows corresponds to external interfaces to other peripherals.}
    \label{fig:ROS2_schema}
\end{figure}

\subsection{Algorithm-level Simulation}
The first abstraction level is exclusively oriented to research and development, as it enables testing the core logic of high-level decision IPP algorithms. 
At this highly abstract level, vehicles are not subject to most real physical or dynamic constraints, so the fleet is considered local in Fig.~\ref{fig:GuadalPlannerScheme}. For example, to move between points they simply “teleport” ignoring temporal constraints. This allows evaluation of the logical and strategic behavior of the algorithms under different conditions, without concern for the performance of the physical platform. 

\begin{figure}[t]
    \centering
    \includegraphics[width=0.48\textwidth]{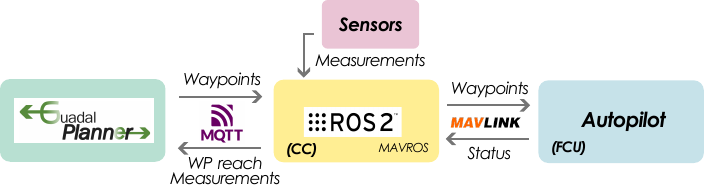}
    \caption{Diagram illustrating the connection flow between the decision-making module (GuadalPlanner), the ROS2 middleware, and the vehicle autopilot.}
    \label{fig:Comms_schema}
\end{figure}

Each simulated vehicle is represented by the fundamental class \textit{Vehicle}, which stores its position, its traveled distance, and performs the movement within the graph-based environment. At each node, the vehicle can take virtual measurements of a synthetic ground truth, stored as node attributes. Although continuous-space formulations is not currently supported in GuadalPlanner, the graph-based abstraction can be defined at higher resolutions through denser discretization, allowing the approximation of continuous environments when required. This choice reflects a trade-off between generality and computational efficiency. Vehicles are grouped and managed by the \textit{Fleet} class, previously mentioned, which coordinates their movements at a higher level, allowing batch execution of commands common to all agents, such as resetting states, collecting data, or calling the movement of the whole fleet. This is the level to validate the logic of the decision algorithm and the interaction of the main classes before adding complexity.

\subsection{Software-in-the-Loop}
The next abstraction level introduces the actual control and communication architecture used in autonomous vehicles, marking the transition from local to remote fleets. At this stage, synthetic movement is replaced by the SITL engine, while environmental data may still be simulated. ArduPilot’s SITL simulator emulates the dynamics and behavior of real autonomous vehicles, allowing the full implementation stack—including ROS2, MAVROS, and planning algorithms—to be tested under conditions that closely approximate real missions, but without requiring physical experiments.

The decision-making process begins with the planning module, which computes the next navigation target based on current mission data and environmental observations. This target point is transmitted via MQTT to the ROS2-based middleware system, which in turn communicates with the vehicle (physical or simulated) using MAVROS as the MAVLink interface. Once the vehicle reaches the specified waypoint, an acknowledgment signal is generated and propagated back through the system: from the SITL vehicle using MAVROS to the middleware, and then to the planning algorithm running in GuadalPlanner. This acknowledgment  triggers the next iteration of the decision cycle, where the planner reevaluates the situation using updated sensor data and mission context.

The middleware is built on ROS2 due to its widespread adoption in robotics research and its native support for modular, distributed, and real-time communication. These features align with the requirements of the proposed architecture, particularly the need for process isolation and the provision of a wide range of modular libraries and tools with standardized interfaces. 
Each process is contained in a structure called a \emph{node}, and they can communicate with each other.
Fig.~\ref{fig:ROS2_schema} shows the modular organization of the middleware nodes responsible for sensor processing, decision-making, communication, and control. 
It is executed on the companion computer (CC), where it acts as the processing core for decision making and system management. 
In addition, it also manages the low-level path planner, responsible for real-time obstacle avoidance by calculating free paths over a high-resolution graph using Dijkstra’s algorithm.

As illustrated in Fig.~\ref{fig:Comms_schema}, communication with the vehicle’s flight controller unit (FCU) is performed through an MQTT-based interface. MQTT is a lightweight publish–subscribe messaging protocol that is well suited for distributed robotic systems operating over limited resources networks, such as those encountered in field deployments using cellular data. An MQTT broker is required to mediate communication between the planning module and the vehicle, and it can be hosted on a local machine or on a remote server accessible through the network. GuadalPlanner may run on the same server or on a separate node, depending on the deployment configuration.
This communication layer enables the planning module to interact with real or simulated vehicles using the same message semantics.
At this abstraction level, GuadalPlanner relies on its remote execution interfaces, which mirror the behavior of their local counterparts while publishing and receiving execution commands via MQTT topics. Execution is synchronized through explicit acknowledgment messages.
The \textit{RemoteFleet} class (see Fig.~\ref{fig:GuadalPlannerScheme}) manages multiple \textit{RemoteVehicle} instances concurrently using Python threads, enabling coordinated multi-agent missions.

\subsection{Real Deployment}
The third level of abstraction transfers the proposed architecture to real operating conditions, using the same planning interfaces used in the algorithm-level simulation and in the SITL. Unlike the second level, this stage interacts directly with an autonomous physical platform, serving as the final phase of deployment.

In this deployment stage, the system leverages the remote interfaces depicted in Fig.~\ref{fig:GuadalPlannerScheme}, following the communication pathway corresponding to the \textit{real world} execution flow. Specifically, high-level planning and monitoring components operate through the remote interfaces, while low-level vehicle execution is handled by the physical platform. This configuration mirrors the interaction pattern used in SITL level, with the only difference being the replacement of the simulated backend by a real vehicle, thereby preserving the same decision-making and communication logic across the entire execution pipeline.
Compatibility at this level requires only a minimal set of system assumptions: an FCU running MAVLink-based firmware (e.g., ArduPilot or PX4), a CC capable of executing ROS2-based middleware and its nodes, mission-specific sensors, and a communication channel for exchanging state and mission data with the MQTT broker. 

For environmental sampling, beyond generating a spatial model from real-time measurements or simulating synthetic ground truths, GuadalPlanner also supports the use of externally provided maps as input data sources. This feature allows, for example, the integration of historical or previously collected environmental data, such as older pollution maps or satellite estimates.

\begin{figure*}[t]
    \centering
    \includegraphics[width=0.85\textwidth]{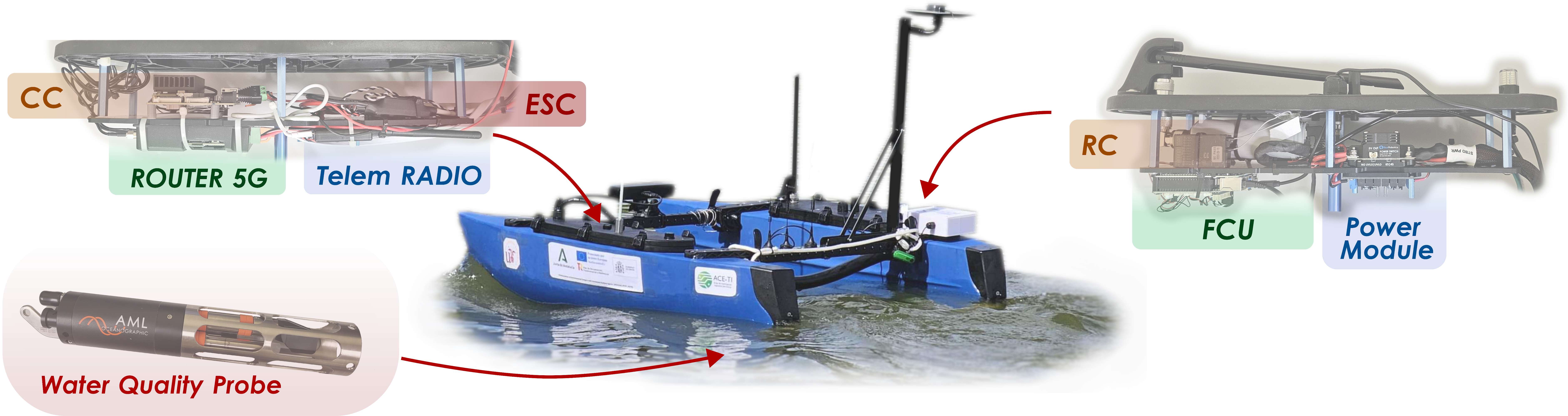}
    \caption{Diagram of the ASV platform illustrating the key onboard components required to support full integration with GuadalPlanner}
    \label{fig:ASV_scheme}
\end{figure*}

\subsection{Software interface use}

To promote reproducibility and facilitate adoption by the research community, the complete implementation of the proposed architecture—including example environments, planning algorithms, and deployment configurations—has been made publicly available in our GitLab repository: \url{https://gitlab.ratatosk.cc/syanes/guadalplanner}. 
This section presents representative examples illustrating how the system can be instantiated and configured to define an IPP task, deploy vehicles, and the transition from simulation to real-world operation.

\begin{pythoncode} 
# Creation of the base graph environment
graphEnv = BaseGraphEnvironment(
            base_matrix=mask_matrix,
            lat_lon_map=lat_lon_matrix)
    
# Initialization of the simulated Ground Truth
graphEnv.fill_graph_attribute('ground_truth', gt)

# Fleet creation
fleet = Fleet(n_vehicles = 2,
            initial_positions = initial_positions,
            graphEnv = graphEnv,
            max_distance = 100.0)
\end{pythoncode}
\refstepcounter{listing}\label{code:fleet_initialization}
\vspace{-8pt}
\renewcommand{\lstlistingname}{Code \thelisting} \noindent\captionof{lstlisting}{Initialization of a GuadalPlanner’s Fleet using a precomputed ground truth for simulation.}
\vspace{20pt}

To run GuadalPlanner in algorithm-level simulation mode, researchers can use the pre-designed environments and algorithms included in the repository, based on the instructions provided therein. However, they can also develop and integrate new strategies without altering the general workflow. A minimal initialization illustrating how to declare a local fleet using GuadalPlanner is shown in Code~\ref{code:fleet_initialization}, serving as a basis for more complex developments.

To run GuadalPlanner on a real vehicle, it is essential that its CC meets the aforementioned requirements. If a SITL simulation environment is chosen, it is necessary to have ArduPilot SITL on the machine where it will be executed, as well as ROS2 and MAVROS. To facilitate the system usage, as detailed in the repository, an automated setup has been provided for experiments at the second abstraction level. 
This setup is a minimal implementation, based on Docker containers, and encapsulates all services required by GuadalPlanner, including the vehicle execution through SITL, the middleware nodes packaged within a preconfigured ROS2 environment, MAVROS as intermediary, and the MQTT communication infrastructure.
When executed, it initializes the autopilot, arms the vehicle, and sets it to autonomous guidance mode, resulting in a system that is ready to receive planning commands from any IPP algorithm implemented within GuadalPlanner. The MQTT broker is deployed locally by default, although an external server can be selected through simple configuration changes.
The encapsulation of all execution dependencies behind a single entry point allows researchers to focus on the development and evaluation of IPP algorithms rather than on system integration.

With the Docker containers running, it is possible to launch one of the available IPP algorithms in remote mode to initiate a mission execution process. Code~\ref{code:remote_initialization} illustrates the declaration of the fleet in this remote configuration. The implementation is similar to the local simulation case, with the key difference being that the MQTT parameters must be specified during the fleet initialization to enable communication with the MQTT broker.

\begin{pythoncode} 
# Remote fleet creation
fleet = RemoteFleet(n_vehicles = 2,
        initial_positions = initial_positions,
        graphEnv = graphEnv,
        max_distance = 100.0,
        mqtt_comm_params = mqtt_comm_params)
\end{pythoncode}
\refstepcounter{listing}\label{code:remote_initialization}
\vspace{-8pt}
\renewcommand{\lstlistingname}{Code \thelisting} \noindent\captionof{lstlisting}{Initialization of a GuadalPlanner’s Remote Fleet. MQTT parameters are now required to communicate with the broker.}
\vspace{20pt}

Finally, a basic main loop of the fleet movement and measurement process is presented in Code~\ref{code:movement}, which works identically for all three levels of abstraction. Target nodes are provided by the IPP algorithm under test. It iteratively receives new measurements from the environment and selects the next target node where it expects the greatest value gain. The difference in the behavior of taking measurements depends on the type of fleet that has been instantiated. If real vehicles are being used, they will take measurements directly from sensors and send them. However, it is also possible to set up a real fleet to send synthetic measurements if desired, e.g., for controlled testing without relying on the physical environment. In simulated environments, measurements are generated synthetically. This uniformity in the main process ensures that the algorithm is completely detached.

\begin{pythoncode} 
# Command movement to target nodes until reached 
while not all(reached.values()): 
    reached, dones = fleet.move(target_nodes) 
# Take a measurement 
measurements = fleet.take_measurement() 
\end{pythoncode}
\refstepcounter{listing}\label{code:movement}
\vspace{-8pt}
\renewcommand{\lstlistingname}{Code \thelisting} \noindent\captionof{lstlisting}{Example of movement command for a fleet and how to take measurements.}
\vspace{20pt}

The code snippets in this subsection have been extracted and adapted for their correct display from the \textit{examples} folder found in the GuadalPlanner repository. This is intended to illustrate the usability of the system and its support for exchanging environments and IPP algorithms.

\section{Experiments}

To evaluate the proposed architecture, a series of experiments were conducted across the three defined levels of abstraction: high-level algorithm simulation, SITL-based control integration, and full deployment on a real ASV. In addition, the physical description of the ASV employed for testing is intended to provide a practical reference for future developers, showing the necessary components to build autonomous platforms compatible with GuadalPlanner. The platform used is detailed in Fig.~\ref{fig:ASV_scheme}. Its onboard architecture consists of two components: an NVIDIA Orin NX as the CC, in charge of managing high-level operations and communications through ROS2, and the Blue Robotics Navigator autopilot as the FCU, in charge of all low-level control tasks of the vehicle. The autopilot, running ArduPilot, interacts directly with critical hardware such as GPS, IMU, ESC thrusters, and RC input. It is responsible for managing low-level control tasks such as movement control, speed regulation, and waypoint navigation.

The vehicle communicates with a remote self-hosted server through a 4G mobile Internet connection, using the MQTT protocol. This server is equipped with high-performance GPUs and multi-core processors, enabling complex computational processing if necessary for demanding decision-making algorithms. It manages key tasks such as maintaining the MQTT broker, storing and processing monitoring data, and generating movement policies for the vehicle.

For the experiments performed, several representative mission scenarios were designed. In the \textit{gpBasedEnv} (see Fig.~\ref{fig:GuadalPlannerScheme}), the vehicle collects samples of a specific water quality parameter (e.g., turbidity, temperature, or pH) with the ASV at multiple locations within a lake area. As seen in works such as \cite{peralta2023water,jaraten2024aquafel}, the objective is to construct a spatial model of the parameter distribution using Gaussian process (GP) regression, a probabilistic method that estimates the parameter value at unsampled locations while providing a measure of uncertainty. 

Another test scenario, \textit{trashCleanEnv}, consists of a plastic waste collection environment based on \cite{barrionuevo2025optimizing}. In this environment, the ASV is tasked to locate and collect floating waste within the predefined water area. Floating debris are randomly distributed on the surface, and the vehicles are equipped with a simulated detection system that identifies these objects within a limited vision radius around their current position. When vehicles visit a node, they automatically pick up the waste items present.

Finally, \textit{OilSpillEnv}, an environment designed to simulate oil spill response scenarios, is also introduced based on \cite{casado2025variational}. In this case, the vehicle must detect and delimit contaminated areas using sensors that measure the concentration of contaminants in the water. As the vehicle moves, it collects samples that feed a probabilistic model of the spill localization, allowing the system to plan trajectories that maximize information coverage. These environment proposals provide diverse and realistic scenarios that enable the evaluation of IPP algorithms within the developed experimental architecture.

Due to the physical limitations and actual capabilities of the ASV platform, real-world tests have been performed in the environment designed to obtain a model based on GPs.
The physical vehicle is equipped with probes capable of measuring water quality parameters, allowing validation of system behavior with real data. In contrast, the other environment proposals, requiring specific sensing and actuation systems not available on the current vehicle, have been validated in simulation.

\begin{figure}[t]
    \centering
    \includegraphics[width=0.48\textwidth]{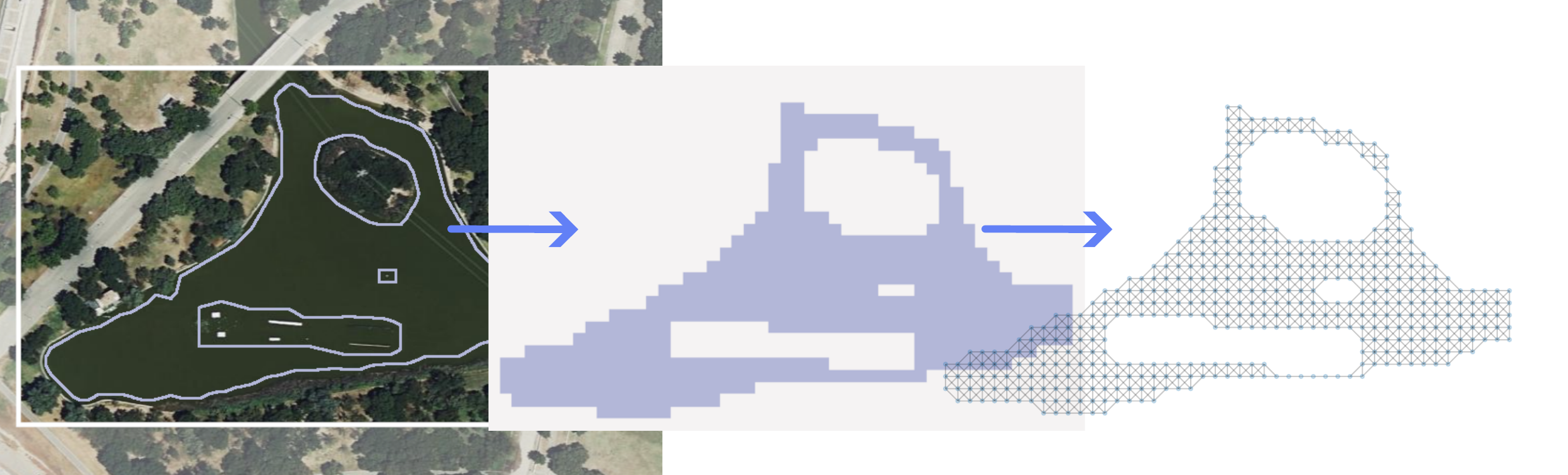}
    \caption{Discretization of the scenario used during the experiments. Left:
Satellite image of lake environment. Middel: Discretized grid representation.
Right: Graph structure derived from the grid.}
    \label{fig:scenario_discretization}
\end{figure}

A common feature of works employing IPP algorithms such as \cite{yanes2024deep,barrionuevo2025optimizing} is the discretization of a real scenario map. The scenario used for both simulation and real-world tests conducted is the Lake Mayor in Alamillo Park (Sevilla, Spain), as this is where the actual vehicle will be deployed. To this end, based on satellite images, a discretization of the location has been made using a resolution of approximately $25 \,\text{m}^2$ per cell. The resulting grid (shown in Fig.~\ref{fig:scenario_discretization}) consists of a $30 \times 49$ mesh, yielding a total of 351 visitable nodes. The resolution of scenario discretizations made for GuadalPlanner use can be adjusted according to the needs of the problem.

The algorithms implemented with the system are three representative autonomous planning approaches, each with a different decision-making strategy, and specially designed for the environment based on GPs. The \emph{greedy} algorithm selects at each step the neighboring node with the highest uncertainty in the GP model (i.e., the highest estimated standard deviation). Its objective is to quickly reduce the uncertainty of the environment, prioritizing poorly explored areas. The \emph{expected improvement} method combines uncertainty and expected value of the sample, selecting points that not only present high variability but also a high probability of improving the overall estimate, balancing exploration and exploitation as in Bayesian optimization. Alternatively, the \emph{flooding} algorithm implements a systematic coverage strategy, in which the vehicle attempts to visit all accessible nodes in an ordered pattern, without prioritizing information or uncertainty. Although not reactive in terms of information acquisition, it serves as a baseline for comparison with informative approaches.

To support these varied planning strategies, the system includes two operational modes, each suited to different algorithmic needs. In \emph{sequential mode}, the model is updated after every movement step, and the system returns control after each action. The vehicle navigates the grid by choosing among the eight cardinal and intercardinal directions, moving to adjacent nodes if available. After each move, a new sample is taken, the GP is updated, and the algorithm selects the next node. This mode is ideal for algorithms such as greedy that require step-by-step decision-making. In contrast, the \emph{target mode} is designed for strategies such as expected improvement, where each vehicle is assigned a distant target node, and the model is updated only once all vehicles have reached their goals. Measurements are still taken at intermediate nodes along the path, allowing data to be collected continuously during the mission. This mode is better suited for strategies that plan over longer horizons while still benefiting from incremental information acquisition along the way.

\subsection{Algorithm over Simulation}
The first set of tests corresponds to the highest level of abstraction, where the focus is on validating the algorithm and the simulation environment. In this simplified setup, vehicles can instantly move between sampling points, without simulating physical dynamics or control. The goal is to test that the implemented layers, the environments, and the IPP algorithms work correctly and interact as expected. This serves as a first step before moving to more realistic scenarios.

\begin{figure}[t]
    \centering
    \includegraphics[width=0.48\textwidth]{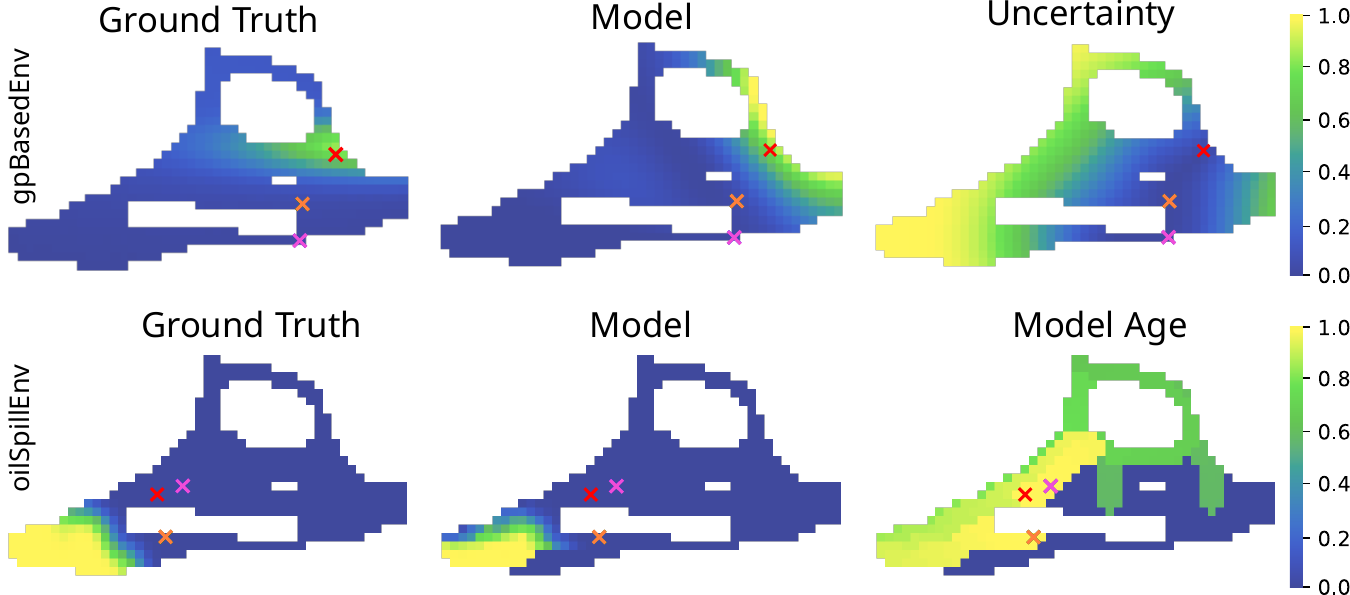}
    \caption{Visualizations of the algorithms running in simulation mode in two
environments with three vehicles in a given step. Vehicle positions are marked
by crosses.}
    \label{fig:Envs_3veh}
\end{figure}

At this level, multi-vehicle coordination has been tested within several environments using a fleet of three simulated vehicles operating concurrently. In Fig.~\ref{fig:Envs_3veh}, this multi-agent operation is illustrated within representative scenarios, where the vehicles collaboratively explore the environment, update the model in real time, and adjust their trajectories based on the evolving information landscape. Visual examples of the simulator running in two different environments are presented. At the top, a frame of the \textit{gpBasedEnv} environment is shown, where three simulated vehicles collaborate by collecting samples from a synthetic ground truth of an environmental parameter. As the ASVs progress, the samples are integrated into a probabilistic model through a GP, which performs the estimation with its corresponding uncertainty. In contrast, the \textit{oilSpillEnv} environment, shown below in Fig.~\ref{fig:Envs_3veh}, implements a different logic: the vehicles do not use statistical models, but directly update the map with the observations they detect within their view radius, overwriting the state of the environment in a deterministic way. Both examples show the versatility of the architecture to integrate different types of decision models, as well as its correct behavior.

\subsection{Algorithm over SITL}
In this second experiment, the system is tested using the ArduPilot SITL simulator, which emulates the physical dynamics of the ASV. The same test scenario and algorithmic structure developed for the previous test is reused, but now the motion commands are sent through MAVROS and MAVLink to a simulated vehicle that must pseudophysically navigate the environment. Therefore, the class used is \textit{RemoteFleet}, which handles real-time position updates and sensor readings via MQTT topics. In this level, the same environments are tested as in the previous one. Now the objective is to validate the ROS2 middleware integration and the communications architecture essential for real deployments, as well as its fusion with the decision algorithms. 

\begin{figure}[t]
    \centering
    \includegraphics[width=0.9\linewidth]{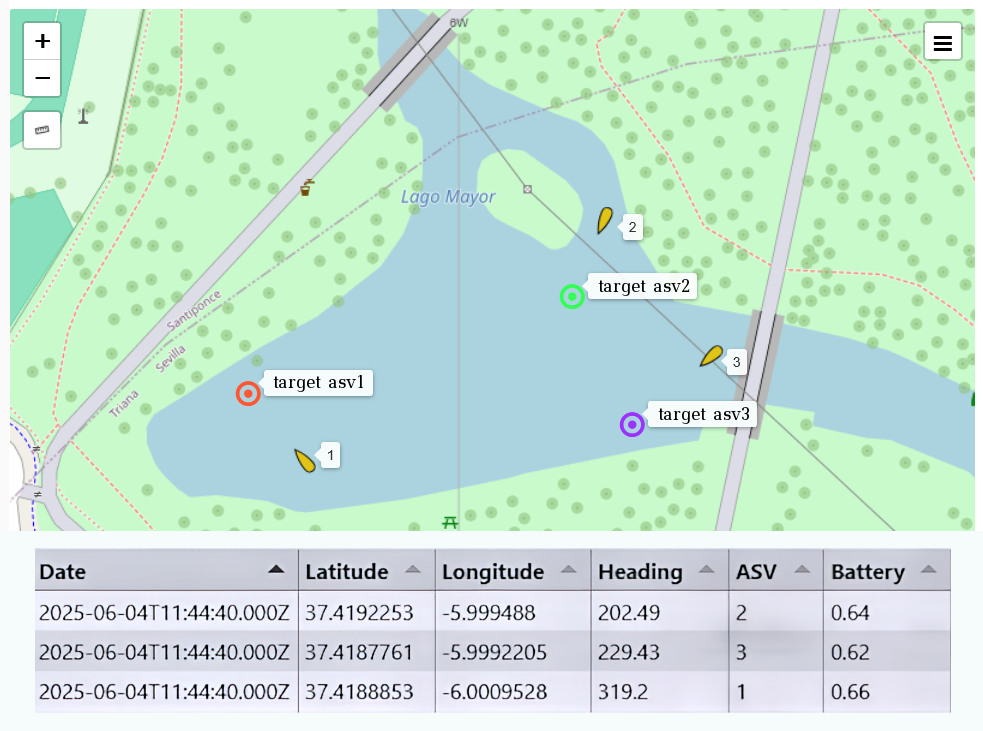}
    \caption{Visualization of the fleet management web interface during a SITL experiment. It runs on the
central server using Node-RED.}
    \label{fig:nodered}
\end{figure}

In Fig.~\ref{fig:nodered}, a remote web interface designed for mission supervision and monitoring is displayed during a SITL mission, showing the simulated GPS positions of the vehicles, the assigned target waypoint for each of them, as well as the real-time storage of collected data in the database.

\subsection{Algorithm over Real World}
In the third and definitive test scenario, the complete system is deployed on a physical ASV platform, operating in a real aquatic environment. In this experiment, instead of using a synthetically generated contamination map, the vehicle conducts a real mission to sample the conductivity parameter of water quality throughout the designated area. The ASV’s onboard sensors collect data, which are transmitted in real time via a 4G IoT connection using the MQTT protocol to the central computation server.

The control software, IPP algorithm, and monitoring systems are executed in the same way as in the SITL simulation test, given the modularity of the system, as well as the communication and command protocols. The algorithms used for this mission are the same as the ones tested in simulation: greedy, expected improvement based on GPs, and flooding. After each sampling step, the server updates the probabilistic model of the environment and generates a new waypoint to be reached by the ASV. The goal for IPP is to progressively reduce the uncertainty of the model by planning movements to the most informative locations, i.e., those with high expected variance. In contrast, the flooding algorithm has a fixed pattern regardless of the data obtained.

\begin{figure}[t]
    \centering
    \includegraphics[width=0.48\textwidth]{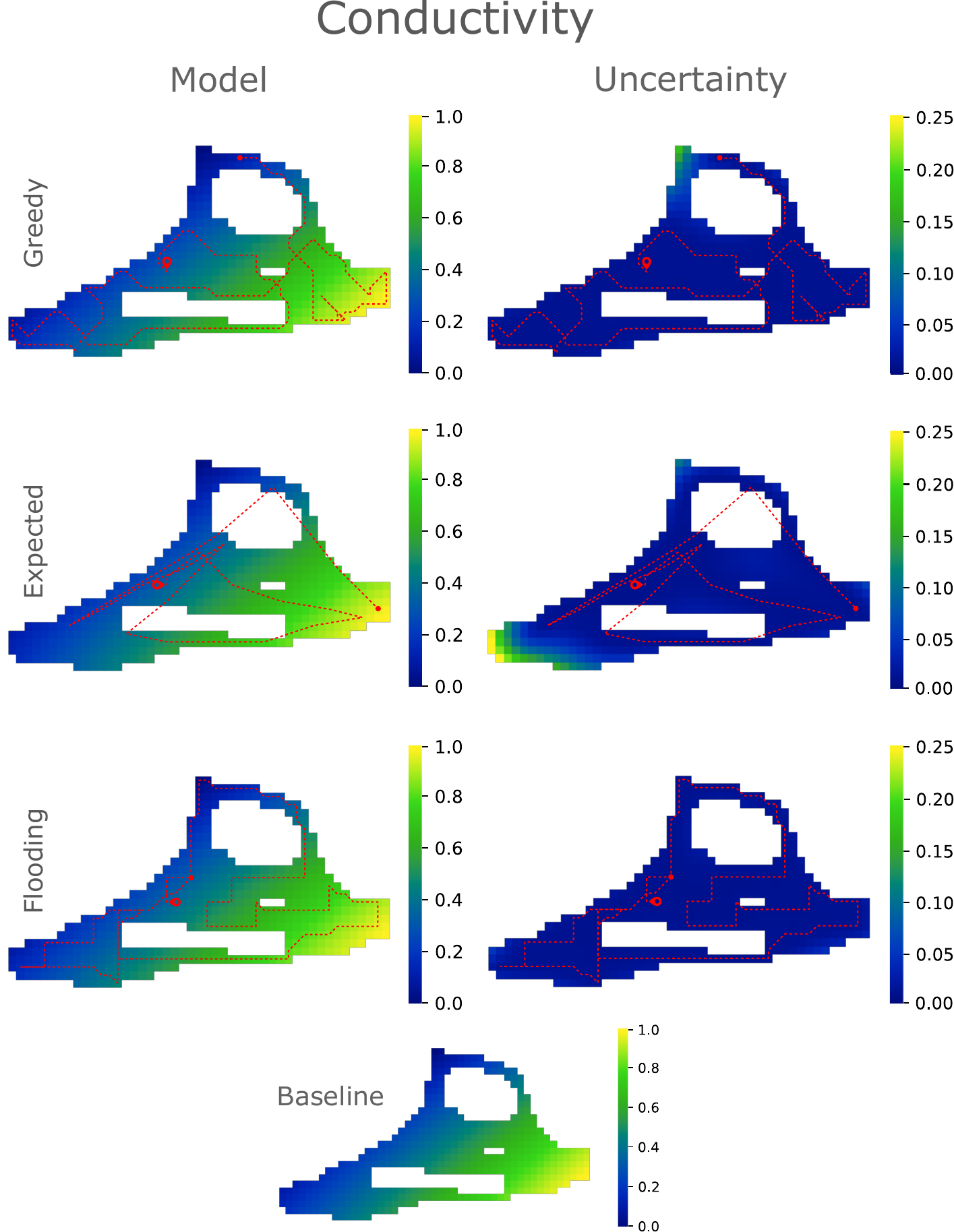}
    \caption{Final results of the GP regression during real world experiments for each tested algorithm are shown. The left column displays the estimated mean distribution, and the right column shows the uncertainty (standard deviation). Target points are joined by straight lines.}
    \label{fig:Results_with_GTmini}
\end{figure}

Despite the logical differences versus simulation, such as GPS noise, water currents, or communication latency, the system remains operational and reactive. The starting point is the same for all algorithms, a central node of the map. In each mission, the vehicle is allowed to travel a maximum distance of 925 meters, defined as an operational constraint to manage energy consumption. This distance is a parameter of the environment that can be adjusted according to the problem requirements. To maintain control and operational safety, the vehicle’s maximum speed is currently limited to approximately 0.5 m/s. Given this speed limitation, a full mission of this magnitude can last up to around 30–35 minutes, depending on path complexity and operational mode selected (sequential or target).

The results obtained by the algorithms in the GP regression environment can be seen in Fig.~\ref{fig:Results_with_GTmini}, which shows the final distribution estimated by the GP and the associated uncertainty. At the bottom, a uniform coverage using a lawn mower pattern that systematically visits all nodes is presented as baseline. These results illustrate how each strategy explores the environment differently and impacts the quality of the reconstructed model. Despite this, all the models obtained are very similar, suggesting that, in this particular scenario, the density and distribution of the points visited were sufficient to adequately capture the structure of the conductivity parameter distribution, regardless of the strategy followed.

\begin{figure}[t]
    \centering
    \includegraphics[width=0.48\textwidth]{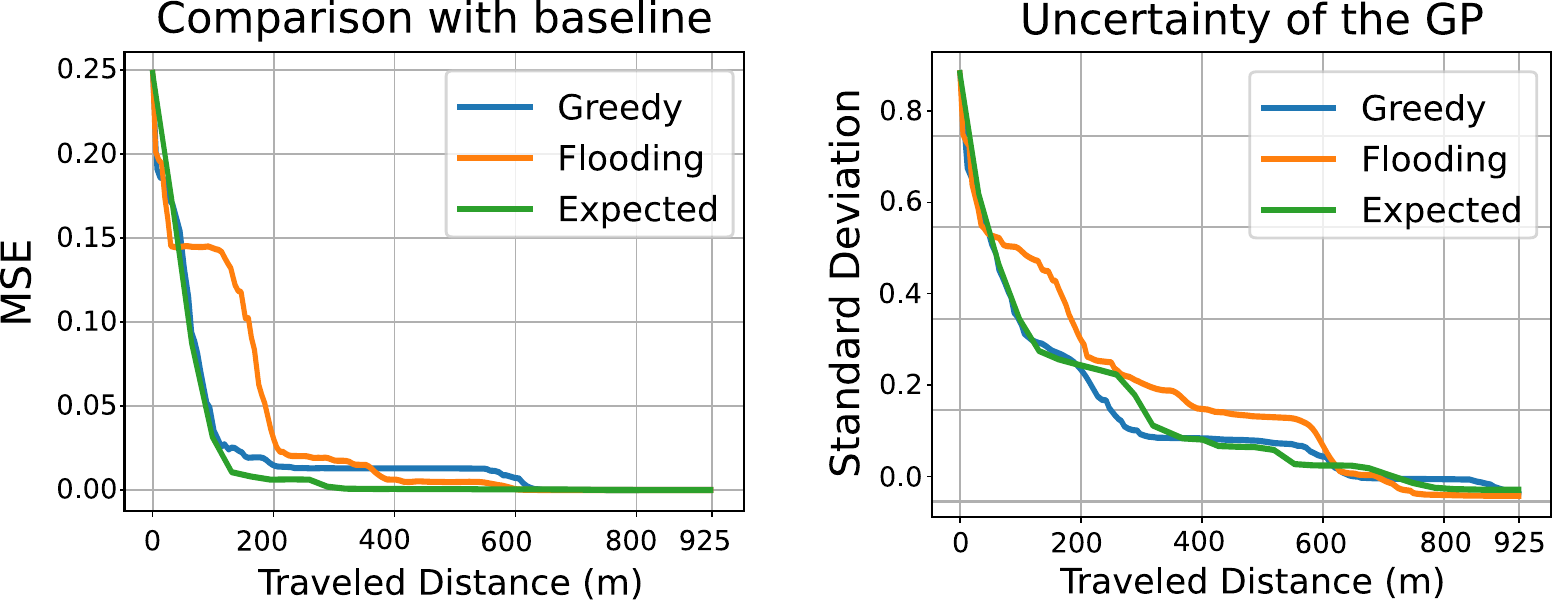}
    \caption{Example metrics during the evolution of the missions.}
    \label{fig:metrics}
\end{figure}

This is evidenced in Fig.~\ref{fig:metrics}, where examples of performance metrics are calculated during the evolution of the mission. The mean squared error (MSE) compared to the baseline shows that all algorithms eventually converge to a model with very low error. The same occurs with the uncertainty of the GP regression, but with a slower decrease, especially for the flooding algorithm. The greedy algorithm tends to focus sampling in high-uncertainty areas, leading to sharp reductions in local variance. The expected improvement method achieves a more balanced distribution of samples, capturing both uncertainty and informative areas. The flooding approach, despite lacking a sense of informativity, provides complete spatial coverage. These metrics are provided as an example of GuadalPlanner’s functionality, but in each environment they can be modified or extended as needed to suit specific mission requirements.

\begin{figure}[t]
    \centering
    \includegraphics[width=0.4\textwidth]{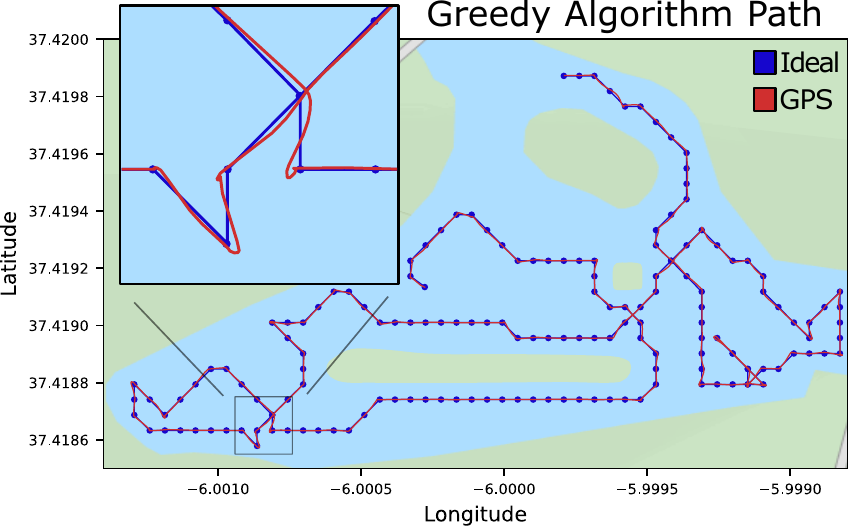}
    \caption{Comparison between the actual GPS trajectory (red) followed by the ASV using the Greedy algorithm and the ideal trajectory (blue).}
    \label{fig:trajectory}
\end{figure}

In Fig.~\ref{fig:trajectory}, the actual trajectory followed by the vehicle with the greedy algorithm is compared to the ideal trajectory defined by the planned target points. The zoomed area highlights deviations in regions of higher navigation complexity, but the vehicle still reaches the target points with sufficient accuracy to maintain the integrity of the sampling process. These small-scale deviations remain within acceptable margins.

\subsection{Limitations}

While the proposed architecture demonstrates consistency, several limitations remain and define clear directions for future work. One such limitation is the reliance on a centralized server for mission coordination and decision-making. This design choice enables global situational awareness and off-board computational resources, which aligns with the centralized nature of many existing IPP approaches \cite{jaraten2024aquafel,peralta2023water,yanes2024deep}. However, it may restrict applicability in scenarios with limited connectivity or where fully distributed autonomy is required. 
However, in operational scenarios, the requirement for a connected server may be replaced by a local ground station which host the communications and planning modules. 
In addition, a different solution could be running GuadalPlanner on each of the vehicles with an MQTT broker on each device. In this way, decentralized algorithms could be implemented.

Another limitation concerns the dependence on MAVLink-based autopilot interfaces, which constrains compatibility to platforms adopting this communication standard. While MAVLink is widely used in aerial, ground, and marine vehicles, supporting alternative or proprietary control stacks would require additional interface layers. Nevertheless, thanks to the open-source philosophy, anyone who needs to can adapt the architecture implementation to suit their project.

Finally, the current implementation of GuadalPlanner is morphologically limited to discrete graph-based environments in two-dimensional spaces. This design choice reflects the predominant representation adopted in the IPP literature \cite{POPOVIC2024104727}, where environments are commonly discretized to enable efficient planning and evaluation. While continuous-space and three-dimensional formulations are not currently implemented, the proposed modular separation is independent of this. Extending the system to continuous or higher-dimensional planning problems is therefore conceptually feasible, but falls outside the scope of this work at the current stage, which focuses on providing a reusable experimental architecture and boilerplate implementation.

\section{Conclusion}

This work has introduces a modular and extensible architecture for Informative Path Planning that covers the complete decision-to-execution pipeline. This architecture has been implemented in Python and named GuadalPlanner.
Rather than introducing new informative planning algorithms, the contribution lies in structuring a coherent execution pipeline that allows IPP methods to be developed, tested, and transferred across different levels of abstraction under consistent assumptions and interfaces. GuadalPlanner is grounded on widely adopted robotics standards such as ROS2, MAVLink, and MQTT.

Through a layered design that separates high-level algorithmic logic from low-level control execution, the proposed architecture enables IPP algorithms to be evaluated in simulation, software-in-the-loop configurations, and real-world platforms without altering their core implementation. 
The experimental results, including deployment in an ASV for monitoring water quality parameters, demonstrate that the architecture can support heterogeneous sensing modalities and information-driven objectives within a unified execution flow. 

Overall, GuadalPlanner aims to contribute a reusable experimental foundation for IPP research, emphasizing methodological consistency and deployability. By lowering the gap between simulation and real-world experimentation, it seeks to support more systematic and transparent evaluation of IPP strategies in autonomous vehicle applications.

\section*{Acknowledgments}
Proyecto PID2024- 158365OB-C21 financiado por MICIU/AEI/10.13039/501100011033 y por FEDER, UE.

\bibliographystyle{ieeetr}
\bibliography{refs}

@inproceedings{khan2014greedy,
  author    = {F. A. Khan and S. A. Khan and D. Turgut and L. B{\"o}l{\"o}ni},
  title     = {Greedy path planning for maximizing value of information in underwater sensor networks},
  booktitle = {39th Annual IEEE Conference on Local Computer Networks Workshops},
  year      = {2014},
  pages     = {610--615},
  publisher = {IEEE}
}

@article{jaraten2024aquafel,
  author    = {M. Jara Ten Kathen and F. Peralta and P. Johnson and I. Jurado Flores and D. G. Reina},
  title     = {Aquafel-PSO: An informative path planning for water resources monitoring using autonomous surface vehicles based on multi-modal PSO and federated learning},
  journal   = {Ocean Engineering},
  volume    = {311},
  pages     = {118787},
  year      = {2024},
  url       = {https://www.sciencedirect.com/science/article/pii/S0029801824021255}
}

@article{peralta2023water,
  author    = {F. Peralta and D. G. Reina and S. L. Toral},
  title     = {Water quality online modeling using multi-objective and multi-agent Bayesian optimization with region partitioning},
  journal   = {Mechatronics},
  volume    = {91},
  pages     = {102953},
  year      = {2023},
  url       = {https://www.sciencedirect.com/science/article/pii/S0957415823000090}
}

@article{yanes2024deep,
  author    = {S. Yanes Luis and D. Shutin and J. Marchal G{\'o}mez and D. G. Reina and S. L. Toral},
  title     = {Deep reinforcement multiagent learning framework for information gathering with local Gaussian processes for water monitoring},
  journal   = {Advanced Intelligent Systems},
  volume    = {6},
  number    = {8},
  pages     = {2300850},
  year      = {2024},
  url       = {https://onlinelibrary.wiley.com/doi/abs/10.1002/aisy.202300850}
}

@inproceedings{figueiredo2009simulate,
  author    = {M. C. Figueiredo and R. J. F. Rossetti and R. A. M. Braga and L. P. Reis},
  title     = {An approach to simulate autonomous vehicles in urban traffic scenarios},
  booktitle = {2009 12th International IEEE Conference on Intelligent Transportation Systems},
  year      = {2009},
  pages     = {1--6},
  publisher = {IEEE}
}

@article{he2019autonomous,
  author    = {X. He and J. R. Bourne and J. A. Steiner and C. Mortensen and K. C. Hoffman and C. J. Dudley and B. Rogers and D. M. Cropek and K. K. Leang},
  title     = {Autonomous chemical-sensing aerial robot for urban/suburban environmental monitoring},
  journal   = {IEEE Systems Journal},
  volume    = {13},
  number    = {3},
  pages     = {3524--3535},
  year      = {2019}
}

@article{liu2022challenges,
  author    = {X. Liu and S. W. Chen and G. V. Nardari and C. Qu and F. Cladera and C. J. Taylor and V. Kumar},
  title     = {Challenges and opportunities for autonomous micro-UAVs in precision agriculture},
  journal   = {IEEE Micro},
  volume    = {42},
  number    = {1},
  pages     = {61--68},
  year      = {2022}
}

@inproceedings{craighead2007survey,
  author    = {J. Craighead and R. Murphy and J. Burke and B. Goldiez},
  title     = {A survey of commercial \& open source unmanned vehicle simulators},
  booktitle = {Proceedings 2007 IEEE International Conference on Robotics and Automation},
  year      = {2007},
  pages     = {852--857},
  publisher = {IEEE}
}

@article{mualla2018comparison,
  author    = {Y. Mualla and W. Bai and S. Galland and C. Nicolle},
  title     = {Comparison of agent-based simulation frameworks for unmanned aerial transportation applications},
  journal   = {Procedia Computer Science},
  volume    = {130},
  pages     = {791--796},
  year      = {2018},
  note      = {The 9th International Conference on Ambient Systems, Networks and Technologies (ANT 2018) / The 8th International Conference on Sustainable Energy Information Technology (SEIT-2018)},
  url       = {https://www.sciencedirect.com/science/article/pii/S187705091830499X}
}

@inproceedings{batista2024deep,
  author    = {L. F. W. Batista and J. Ro and A. Richard and P. Schroepfer and S. Hutchinson and C. Pradalier},
  title     = {A deep reinforcement learning framework and methodology for reducing the sim-to-real gap in ASV navigation},
  booktitle = {2024 IEEE/RSJ International Conference on Intelligent Robots and Systems (IROS)},
  year      = {2024},
  pages     = {1258--1264},
  publisher = {IEEE}
}

@article{kumar2019simulation,
  author    = {G. P. Kumar and B. Sridevi},
  title     = {Simulation of efficient cooperative UAVs using modified PSO algorithm},
  journal   = {WSEAS Transactions on Information Science and Applications},
  volume    = {16},
  pages     = {94--99},
  year      = {2019}
}

@article{barrionuevo2025optimizing,
  author    = {A. M. Barrionuevo and S. Y. Luis and D. G. Reina and S. L. T. Mar{\'i}n},
  title     = {Optimizing plastic waste collection in water bodies using heterogeneous autonomous surface vehicles with deep reinforcement learning},
  journal   = {IEEE Robotics and Automation Letters},
  volume    = {10},
  number    = {5},
  pages     = {4930--4937},
  year      = {2025}
}

@article{casado2025variational,
  author    = {A. Casado-P{\'e}rez and S. Yanes and S. L. Toral and M. Perales-Esteve and D. Guti{\'e}rrez-Reina},
  title     = {Variational autoencoder for the prediction of oil contamination temporal evolution in water environments},
  journal   = {Sensors},
  volume    = {25},
  number    = {6},
  pages     = {1654},
  year      = {2025},
  url       = {https://www.mdpi.com/1424-8220/25/6/1654}
}

@INPROCEEDINGS{IPPwiman,
  author={Wiman, David and Lindgren, David},
  booktitle={2025 11th International Conference on Automation, Robotics, and Applications (ICARA)}, 
  title={Real-Time Capable Informative Path Planning for Autonomous Area Monitoring Using Unmanned Ground Vehicles}, 
  year={2025},
  volume={},
  number={},
  pages={262-267},
  doi={10.1109/ICARA64554.2025.10977641}}

@ARTICLE{IPPyu,
  author={Yu, Ying and Zheng, Huarong and Xu, Wen},
  journal={IEEE Transactions on Systems, Man, and Cybernetics: Systems}, 
  title={Learning and Sampling-Based Informative Path Planning for AUVs in Ocean Current Fields}, 
  year={2025},
  volume={55},
  number={1},
  pages={51-62},
  doi={10.1109/TSMC.2024.3370177}}

@ARTICLE{IPPruckin,
  author={Rückin, Julius and Magistri, Federico and Stachniss, Cyrill and Popović, Marija},
  journal={IEEE Transactions on Robotics}, 
  title={An Informative Path Planning Framework for Active Learning in UAV-Based Semantic Mapping}, 
  year={2023},
  volume={39},
  number={6},
  pages={4279-4296},
  doi={10.1109/TRO.2023.3313811}}

@INPROCEEDINGS{IPPalrashdia,
  author={Al-Rashdia, Hind and Al-Abri, Said and Al-Maashri, Ahmed and Bourdoucen, Hadj},
  booktitle={2025 11th International Conference on Control, Automation and Robotics (ICCAR)}, 
  title={Vision-Based Informative Path Planning for Estimating Oil Spill Area Using Multiple Drones}, 
  year={2025},
  volume={},
  number={},
  pages={283-288},
  doi={10.1109/ICCAR64901.2025.11072942}}

@INPROCEEDINGS{IPPangelyn,
  author={Mercado, Marie Angelyn and Weng, Yang and Maki, Toshihiro},
  booktitle={2025 IEEE Underwater Technology (UT)}, 
  title={Adaptive Informative Path Planning for Lightweight AUV in Coastal Benthic Environments}, 
  year={2025},
  volume={},
  number={},
  pages={1-7},
  doi={10.1109/UT61067.2025.10947450}}

@INPROCEEDINGS{IPPaltahat,
  author={Al-Tahat, Shayma and Al-Bzoor, Manal and Almasaeid, Hisham},
  booktitle={2025 16th International Conference on Information and Communication Systems (ICICS)}, 
  title={Intelligent Informative Path Planning Approach for Autonomous Underwater Vehicles}, 
  year={2025},
  volume={},
  number={},
  pages={1-6},
  doi={10.1109/ICICS65354.2025.11073096}}

@INPROCEEDINGS{IPPkailas,
  author={Kailas, Siva and Deolasee, Srujan and Luo, Wenhao and Kim, Woojun and Sycara, Katia},
  booktitle={2025 IEEE International Conference on Robotics and Automation (ICRA)}, 
  title={Integrating Multi-Robot Adaptive Sampling and Informative Path Planning for Spatiotemporal Natural Environment Prediction}, 
  year={2025},
  volume={},
  number={},
  pages={11413-11419},
  doi={10.1109/ICRA55743.2025.11128099}}

@INPROCEEDINGS{IPPlindgren,
  author={Lindgren, David},
  booktitle={2023 27th International Conference on Methods and Models in Automation and Robotics (MMAR)}, 
  title={Multiple Agent Path Planning for Autonomous Area Monitoring}, 
  year={2023},
  volume={},
  number={},
  pages={22-26},
  doi={10.1109/MMAR58394.2023.10242578}}

@ARTICLE{IPPdutta,
  author={Dutta, Shamak and Wilde, Nils and Smith, Stephen L.},
  journal={IEEE Transactions on Robotics}, 
  title={Informative Path Planning for Active Regression With Gaussian Processes via Sparse Optimization}, 
  year={2025},
  volume={41},
  number={},
  pages={2184-2199},
  doi={10.1109/TRO.2025.3548865}}

@INPROCEEDINGS{IPPyunze,
  author={Hu, Yunze and Chen, Jiaao and Zhou, Kangjie and Gao, Han and Li, Yutong and Liu, Chang},
  booktitle={2023 IEEE 26th International Conference on Intelligent Transportation Systems (ITSC)}, 
  title={Informative Path Planning of Autonomous Vehicle for Parking Occupancy Estimation}, 
  year={2023},
  volume={},
  number={},
  pages={3304-3310},
  doi={10.1109/ITSC57777.2023.10422020}}

@article{ghasemi2024flood,
  title={Flood algorithm (FLA): an efficient inspired meta-heuristic for engineering optimization},
  author={Ghasemi, Mojtaba and Golalipour, Keyvan and Zare, Mohsen and Mirjalili, Seyedali and Trojovsk{\`y}, Pavel and Abualigah, Laith and Hemmati, Rasul},
  journal={The Journal of Supercomputing},
  volume={80},
  number={15},
  pages={22913--23017},
  year={2024},
  publisher={Springer}
}

@article{POPOVIC2024104727,
title = {Learning-based methods for adaptive informative path planning},
journal = {Robotics and Autonomous Systems},
volume = {179},
pages = {104727},
year = {2024},
issn = {0921-8890},
doi = {https://doi.org/10.1016/j.robot.2024.104727},
url = {https://www.sciencedirect.com/science/article/pii/S0921889024001118},
author = {Marija Popović and Joshua Ott and Julius Rückin and Mykel J. Kochenderfer},
}

\vfill

\end{document}